\documentclass[letterpaper, 10 pt, conference]{ieeeconf}  

\IEEEoverridecommandlockouts                              
\usepackage{xcolor}
\usepackage{comment}
\usepackage{graphics} 
\usepackage{epsfig} 
\usepackage{amsmath} 
\usepackage{amssymb}  
\usepackage{mathtools}
\usepackage{marginnote}
\usepackage{hyperref}
\usepackage{graphicx}
\usepackage{cite}
\usepackage{caption}
\usepackage{subcaption}
\usepackage{afterpage}
\usepackage{url}
\usepackage{breakurl}
\hypersetup{breaklinks=true}
\usepackage{xurl}

\title{\LARGE \bf
STRIDE: An Open-Source, Low-Cost, and Versatile Bipedal Robot Platform for Research and Education} 

\author{Yuhao Huang\mbox{*}, Yicheng Zeng\mbox{*}, and Xiaobin Xiong  
\thanks{\mbox{*} The authors contribute equally to this work.}
\thanks{$^1$ This work is supported in part by the University of Wisconsin-Madison Office of the Vice Chancellor for Research and Graduate Education.}
\thanks{The authors are with the Wisconsin Expeditious Legged Locomotion (WELL-Lab) at the University of Wisconsin-Madison.
        {Corresponding to \tt\small xiaobin.xiong@wisc.edu}}%
}

\begin{document}

\newcommand{\Kang}[1]{{\color{blue} #1}}

\newcommand{\block}[1]{\noindent{\textbf{#1}:}}
\newcommand{\emphhh}[1]{{\color{yellow} \textbf{#1}}}
\newcommand{\blockUnderline}[1]{\noindent{\underline{#1}:}}

\maketitle
\thispagestyle{empty}
\pagestyle{empty}



\begin{abstract}
In this paper, we present STRIDE, a \underline{S}imple, \underline{T}errestrial, \underline{R}econfigurable, \underline{I}ntelligent, \underline{D}ynamic, and \underline{E}ducational bipedal platform. STRIDE aims to propel bipedal robotics research and education by providing a cost-effective implementation with step-by-step instructions for building a bipedal robotic platform while providing flexible customizations via a modular and durable design. Moreover, a versatile terrain setup and a quantitative disturbance injection system are augmented to the robot platform to replicate natural terrains and push forces that can be used to evaluate legged locomotion in practical and adversarial scenarios. We demonstrate the functionalities of this platform by realizing an adaptive step-to-step dynamics based walking controller to achieve dynamic walking. Our work with the open-soured implementation shows that STRIDE is a highly versatile and durable platform that can be used in research and education to evaluate locomotion algorithms, mechanical designs, and robust and adaptative controls.\\
\\  
\noindent  Project Repository: \href{https://github.com/well-robotics/STRIDE}{https://github.com/well-robotics/STRIDE}\\
\noindent  Project Video: \href{https://youtu.be/wJkxvUG6msU}{https://youtu.be/wJkxvUG6msU}
\end{abstract}

\section{Introduction}


Bipedal robots are garnering unprecedented attention due to their potential capability to work in human society. Their potential to navigate in human environments and utilize human-sized tools have demonstrated significant value in alleviating people from completing repetitive dull, dear, dirty, and dangerous tasks. Existing platforms such as Boston Dynamics Atlas \cite{BostonDynamics}, Agility Robotics Digit \cite{Agility}, and Disney Research BD-1 \cite{Disney} have already shown their capability in the manufacturing, transportation, and entertainment industries. 

 However, present-day, normal commercial \cite{Agility, BostonDynamics, Disney} and research bipedal robot platforms\cite{AMBER},\cite{RABBIT} are closed-source, mechanically un-configurable, and expensive. For instance, the previously purchasable robot Digit was priced at \$250,000, and the cheaper robot Unitree G1 Edu is priced at over \$30,000 in 2024, making them inaccessible to use in classrooms or even in most research labs, especially in third-world countries. Moreover, the increased complexity of the system reduces its durability; all these robots require a team of engineers for maintenance in research and development. Design modification and customization on commercial robots are also challenging. Consequently, 
 researchers are prone to choose disparate platforms for their bipedal locomotion research \cite{Ames_review, Wensing_review, MPC_review}, which challenge scientific comparisons and systematic evaluations of algorithms on similar robots. These issues can be partly addressed by increasingly realistic simulators. However, simulators also have limitations, particularly in areas such as contact simulation with deformable terrains \cite{simulation}. All these factors hinder the researchers and educators from expediting further contributions to this field. Hence, the need for a unified platform that enables algorithm prototyping and evaluation is critical for fostering scientific progress in bipedal locomotion research.

\begin{figure}[t]
    \centering
    \includegraphics[width=1.0\linewidth]{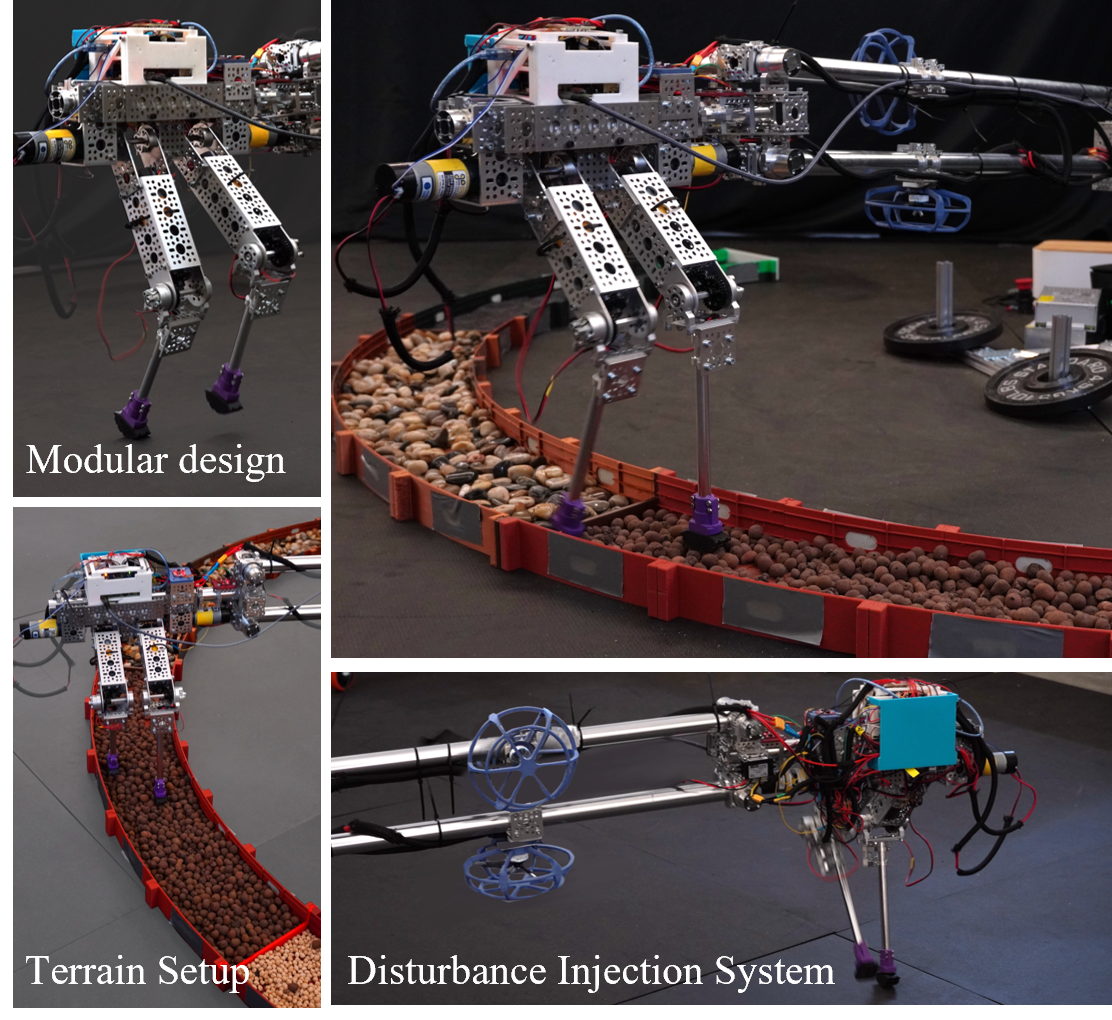}
    \caption{Open-source robot STRIDE.}
    \vspace{-10pt}
    \label{fig:enter-label}
    \vspace{-5pt}
\end{figure} 

Some open-sourced bipedal robot platforms \cite{Hector, Bolt, Poppy, Hoppy, igus} have been developed to address these problems. However, they still encounter challenges in certain aspects. Some of these platforms heavily rely on existing commercial products, limiting access of users to detailed design information and hardware resources; the platforms based on custom-designed metal parts \cite{Hector, saloutos2023design} can require laborious and expensive machining, making them difficult to build and customize. Although these platforms are cheaper than existing commercial ones, they still represent a significant financial burden for educators and individual researchers. An increasing number of open-source platforms are opting for 3D-printed mechanical structures\cite{Bolt, Poppy,igus}, but they often struggle with mechanical durability, particularly when performing locomotion with repeated impact with the ground. These structures also lack mechanical precision, making it difficult to evaluate the performance of the controllers. Therefore, there is still a huge necessity for having low-cost, customizable, and robust bipedal robotic platforms in academia.

In this paper, we propose STRIDE, a planar bipedal robot platform to fulfill this urgent demand for accessible research and educational hardware platforms. The platform is low-cost (below \$2000) and easy to customize without the need of machining. The platform provides dynamic locomotion capabilities, and particularly, it features the following novel functionalities for evaluating legged locomotion:

\begin{enumerate}
    \item Versatile Terrain Setup: A modular terrain setup is provided to enable the evaluation of locomotion over natural terrains. 
    \item Quantitative Disturbance Injection: A set of propellers is utilized to provide measurable push forces to the robot for robustness evaluation of the controllers.
    \item Modular Components for Design Evaluation: A robot design with modular parts allows easy reconfigurations to test design optimization algorithms.
\end{enumerate}
The whole design with software implementation is open-sourced at \cite{git}. The video demonstration of STRIDE can be seen in \cite{video}.  

The rest of this paper is organized as follows. Section \ref{sys design} introduces the overall robot design. Section \ref{control} describes an example of the control implementation for the robot. Section \ref{exp} presents several experiments to demonstrate the performance, robustness, and versatilities of the platform. Finally, Section \ref{conclu} draws a conclusion to our work and points out future directions.



\section{System Design}\label{sys design}


\block{Overview} The design of STRIDE aims to be low-cost, reconfigurable, durable, and mechanically strong. We also exert effort to lower the assembly difficulty, making it straightforward even for novices. The mechanical design primarily consists of off-the-shelf metal components from goBILDA and a small number of custom-designed 3D-printed parts, ensuring both strength and cost-efficiency. Its electrical components include widely accessible units such as Arduino and Raspberry Pi, further enhancing the affordability and accessibility of the robot. Its software consists of commonly used tools such as ROS2 \cite{ROS2}, MuJoCo \cite{Mujoco}, and FROST \cite{FROST}, making the system modular and expandable. 

\block{Education Potentials} A wide range of robotic knowledge can be gained from building and implementing controllers on STRIDE, making it a versatile platform for teaching various topics, including but not limited to robot design, control theory and engineering, and ROS2. Specialized topics in legged robotics, such as floating-based system modeling, underactuated robotics theory, and trajectory planning and optimal control, can also be instrumented on this system. The open-source software repository serves as a template for implementing software on legged robots, further enhancing its educational value.

\block{Research Potentials} Apart from the basic functionality of the robot, STRIDE is also designed with significant research potential. The control software is highly generalizable, allowing for the implementation, evaluation and comparisons of different control \cite{gibson2021terrain,he2024cdm,gong2022zero, xiong2021robust, xiang2024adaptive, acosta2023bipedal} and learning-based algorithms \cite{li2024reinforcement, castillo2023template, van2024revisiting} for synthesizing bipedal walking. The modular environment setup can simulate outdoor conditions such as wind disturbances and rough terrain. Additionally, the modular design of the robot allows research evaluation of hardware design, such as robot design optimization. This can be conveniently implemented by using modular metal and 3D-printed components to adjust both the geometric and dynamic properties of linkages.

\subsection{Mechanical Components}

\begin{figure}[t]
    \centering
    \setlength{\fboxsep}{0pt}
    \colorbox{white}{\includegraphics[width=1.0\linewidth]{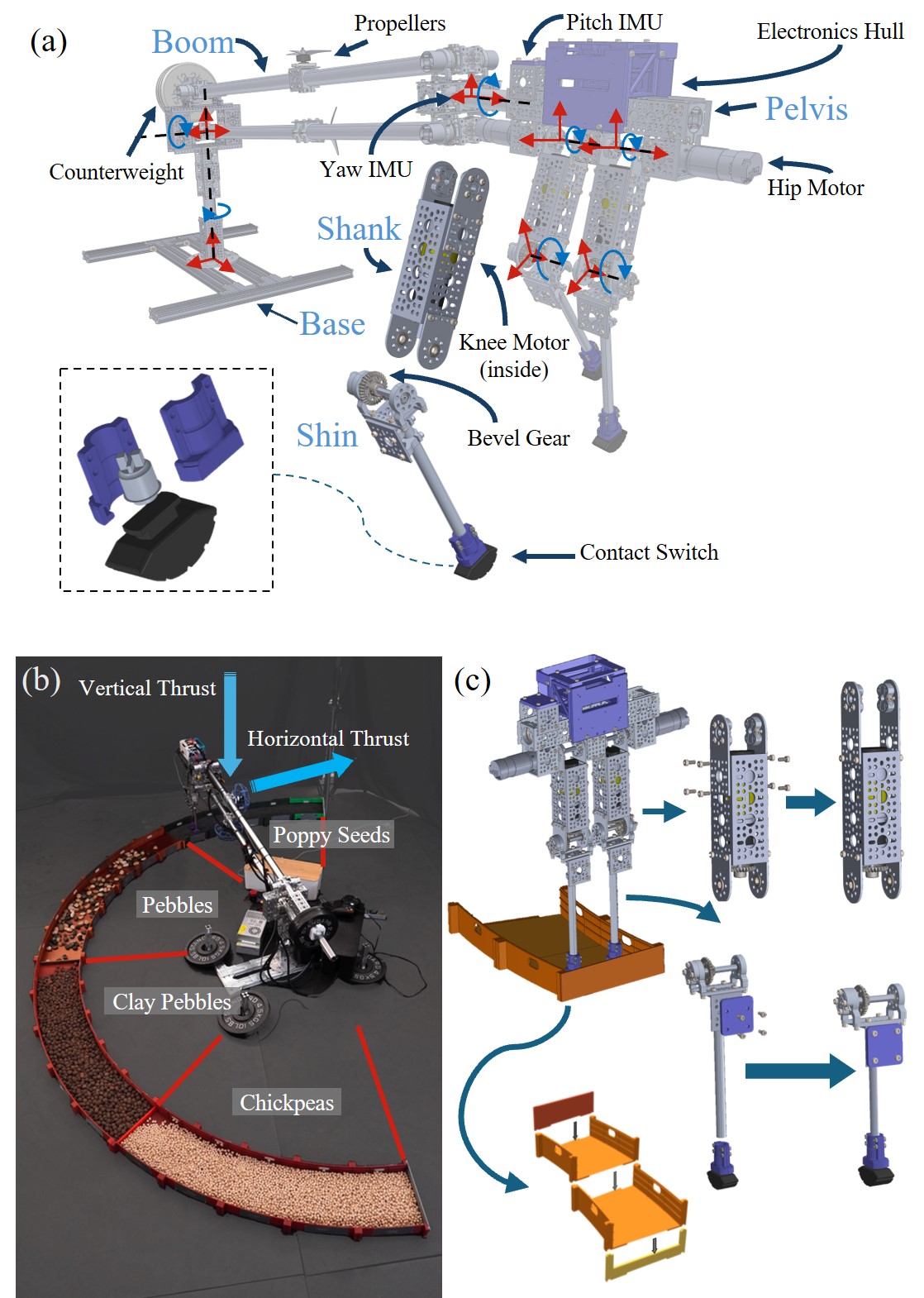}}
    \caption{(a) Mechanical design of STRIDE. (b) System setup with rough terrains and disturbance injection system. (c) Demonstration of modularity and easy customization.}
    \vspace{-10pt}
    \label{fig: mechanical design}
\end{figure}

Fig.~\ref{fig: mechanical design} (a) shows the overall breakdown of the mechanical design of the robot. The system consists of a planar five-link bipedal robot and a parallel four-bar linkage boom that is connected to a fixed base. This configuration resembles the classical planar bipedal platforms such as RABBIT \cite{RABBIT}, AMBER \cite{AMBER}, and ERNIE \cite{ERNIE}. To ensure the robot has predominantly planar motion, the four-bar linkage boom constrains the robot in its sagittal plane and prevents lateral motion. Counterweights on the opposite end of the boom balance its gravitational forces, thereby minimizing the impact of boom dynamics to the robot dynamics. Two propellers are mounted on the four-bar linkages orthogonally as shown in Fig. \ref{fig: mechanical design} (b), which can provide accurate and measurable disturbance forces to the floating-base of the robot. A low-cost and versatile terrain setup is built for evaluating locomotion over natural terrains. All the components are modular and self-contained so that one can easily repurpose them for different purposes. 

\blockUnderline{Robot Design} Each leg of the robot has two actuated joints: the hip motor is placed near the torso, and the knee motor is placed inside the thigh. Bevel gears are employed to transmit power from knee motors to the knee joints. This configuration reduces the inertia of the legs and enables rapid control of leg swinging. The brushed DC actuators are selected to meet both torque and angular speed criteria while maintaining a minimal gear ratio to reduce reflected inertia. All the actuators are from goBILDA (84 RPM, 93.6 kg${}\cdot{}$cm); it offers actuators with the same form factor but with different gear ratios and output shafts, which allows custom selections of actuators if needed. Contact detection is achieved through cheap contact switches with 3D-printed curved feet that can tolerate repetitive impacts with the ground. 

\blockUnderline{Disturbance Injection} The propellers offer a more precise and measurable approach to inject disturbances to the robot, compared to the commonly seen method where a human operator simply pushes with unknown magnitudes \cite{HLIP_TRO}. The two propellers are installed orthogonally to provide disturbance forces to any direction in the plane, which allows thorough evaluations of the robustness of controllers.

\blockUnderline{Terrain Setup} The entire terrain setup consists of 3D-printed modular Lego-like structures that can be stacked up piece by piece. It can contain rough materials such as poppy seeds, garbanzo beans, cobblestones, and rocks to emulate natural terrains. In this design, we choose poppy seeds, pebbles, clay pebbles, and chickpeas as the filling materials. Poppy seeds are granular and soft, while pebbles are relatively large and slippery. Clay pebbles are of the same size as pebbles but with rougher surfaces, whereas chickpeas have similar roughness with smaller diameters. The cost of the terrain is estimated at \$200 for one full circular track.


\subsection{Electronic Components}
\begin{figure}[t]
    \centering
    \includegraphics[width=1.0\linewidth]{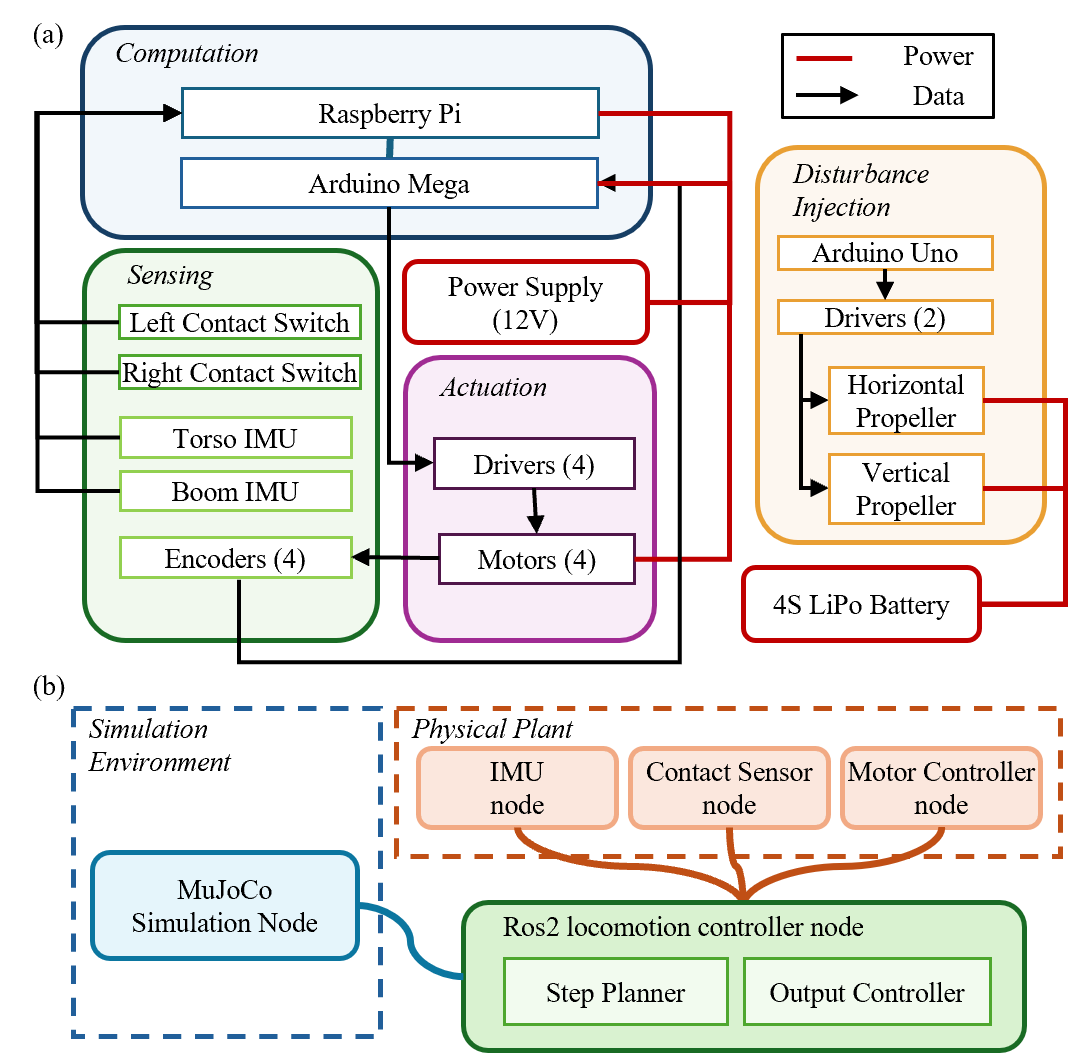}
    \caption{(a) Electrical components and wiring topology. (b) Software architecture, in which the ROS2 controller node can be customized by the user.}
    \vspace{-10pt}
    \label{fig:Electronics}
\end{figure}

The electronic system is designed to be modular and easily reconfigurable for different research purposes. Fig. \ref{fig:Electronics} (a) depicts the electrical components and communication protocols used in the robot. The whole electrical system has a user computer (Raspberry Pi 4b+) for implementing robot control and sensor processing and a low-level microcontroller (Arduino Mega 2560 R3) that has implemented motor control algorithm. Communication between the Raspberry Pi and the Arduino is implemented through a USB. These components are chosen based on the balance of cost and performance, ensuring that the robot has sufficient computation capacity. Additionally, all electronic components are mounted on different layers of a housing unit, allowing for easy inspection and substitution of individual components. This design simplifies hands-on assembly and eases the debugging of the electronic system.

Two IMUs (SparkFun ICM-20948 9DoF IMU) and two contact switches are connected to the Raspberry Pi for gait synthesis. One IMU is placed on the robot pelvis to measure the pelvis angle and angular velocity, while another is mounted at the end of the four-bar linkage, as shown in Fig.~\ref{fig: mechanical design} (a), to measure the rotational velocity of the circular motion. Compared to an encoder, an IMU is easier to assemble while providing sufficiently accurate measurements. Two contact switches are placed on the foot to measure the foot-ground contact. The Arduino Mega controls the brushed DC actuators by sending Pulse Width Modulation (PWM) signals to the Motor Drivers (VNH5019 12A, 5.5-24V); the Arduino also reads the pulse counts from the encoders ($1993.6$ pulses per revolution) built inside the actuators. The motor drivers, Arduino, and Raspberry Pi are powered by a 12V power supply (MEISHILE 12V 40A 480W). 

The disturbance injection system includes two propellers, two brushless DC motors (T-Motor V2207 2550KV), and two motor controllers (Tiger Motor F35A) connected to an Arduino Uno. The Arduino receives commands from a joystick via a USB communication, and it sends out control signals (DShot300) to the motor driver to control the motors. This setup allows users to provide both continuously changing and fixed amount disturbance forces to the robot based on their needs.

The entire electronic setup is easily reconfigurable. The disturbance injection system can share the same Arduino as the motor control system, eliminating the need for an additional Arduino Uno. Additionally, the joystick can be connected to the Raspberry Pi, allowing it to control both the robot and the disturbance injection system if needed.

\subsection{Software}
The software architecture of the robot is designed to be efficient, hardware-friendly, modular, and expandable. The code is developed using state-of-the-art software environment ROS2, making it suitable for education. Hardware interfaces and controllers are implemented in \texttt{C++} and leverage the \texttt{Eigen} library for linear algebras, ensuring efficient computation and compatibility with various hardware configurations. Along with these codes, a separate software package for simulation and control is provided in MATLAB, allowing users to quickly validate their research works and visualize results through straightforward simulation and animation.

 Fig. \ref{fig:Electronics} (b) depicts the overall architecture of our software package that has two major components: hardware interfaces and the controller. The hardware interfaces process raw sensor readings to publish them as appropriate message types for walking control. The controller subscribes to sensor messages, plans desired motion, and solves for the motor control commands which are published as ROS2 topics. An example of the controller implementation of walking is described in section IV. To support custom implementations of controllers, we also provide all the kinematics and dynamics libraries of the robot in the software package. 

In parallel to the hardware nodes, a simulation node using MuJoCo \cite{Mujoco} has been implemented. In this simulation, the robot is constrained in its sagittal plane virtually to mimic the designed planar motion. Sensor data in the simulation are formatted and directly sent to the controller node. In this way, the same controller node can be evaluated in both simulation and hardware. This architecture minimizes the sim-to-real transfer iterations and ensures rapid implementation debugging in simulation before deployment on the hardware, which minimizes the potential risks of damaging the hardware with buggy controls.  It also facilitates debugging in various scenarios by replicating real-world conditions, e.g., observing the effects of excessive or insufficient feed-forward torque, the impact of sensor noise, and variation in locomotion performance w.r.t. controller parameters.

\subsection{Modularity and Customization}

The robot is designed to facilitate the customization of both hardware and software. goBILDA provides metal parts of various lengths and actuators with different gear ratios. Additionally, the mass and inertia properties of the robot can be easily modified by attaching 3D-printed parts to its body. The design approach simplifies modifications on both the kinematic and dynamic properties of the robot. An illustration of its modularity is illustrated in Fig. \ref{fig: mechanical design} (c). To change the link lengths or add masses to the links, one only needs to remove a few screws, swap the parts, and tighten the screws again. The terrain setup is designed for easy assembly and disassembly, with flexible adjustments in its length and the terrain it contains. As shown in Fig. \ref{fig: mechanical design} (c), terrain blocks can be easily stacked or removed. A connector block can be inserted to combine two terrain blocks while separating different terrain materials if needed. Moreover, the electrical components can be readily substituted for any appropriate ones due to the modular design of the electronic hull. The main computer can be swapped to other single-board computers such as an Intel UpBoard or NUC, and the motor controllers can be upgraded to advanced motor drivers that are capable of performing current regulation.

 The software, both the simulation and hardware, are written as ROS2 packages, and thus can be easily customized for different educational and research purposes. \texttt{C++} enables highly efficient computation and direct integration with ROS2, while ROS2 allows for seamless integration with other open-source packages in terms of control, planning, and estimation. For example, a quadratic programming (QP) based controller from \cite{HLIP_TRO} can be easily integrated into the control implementation example, by substituting the Inverse Kinematics (IK) and PD-based controller. The modular design of both the software and hardware will facilitate the rapid evaluation of novel controls and hardware designs.

\section{Control Implementation example} \label{control}
This section describes a detailed control implementation example of STRIDE. The Step-to-Step Dynamics (S2S) based walking controller \cite{HLIP_TRO, Ad_HLIP} is used to synthesize the desired walking behaviors due to its general formulation of walking that can be used for robots with different hardware configurations, its robustness and adaptation to disturbances, and its straightforward implementation.
\begin{figure}[t]
    \centering
    \includegraphics[width=1.0\linewidth]{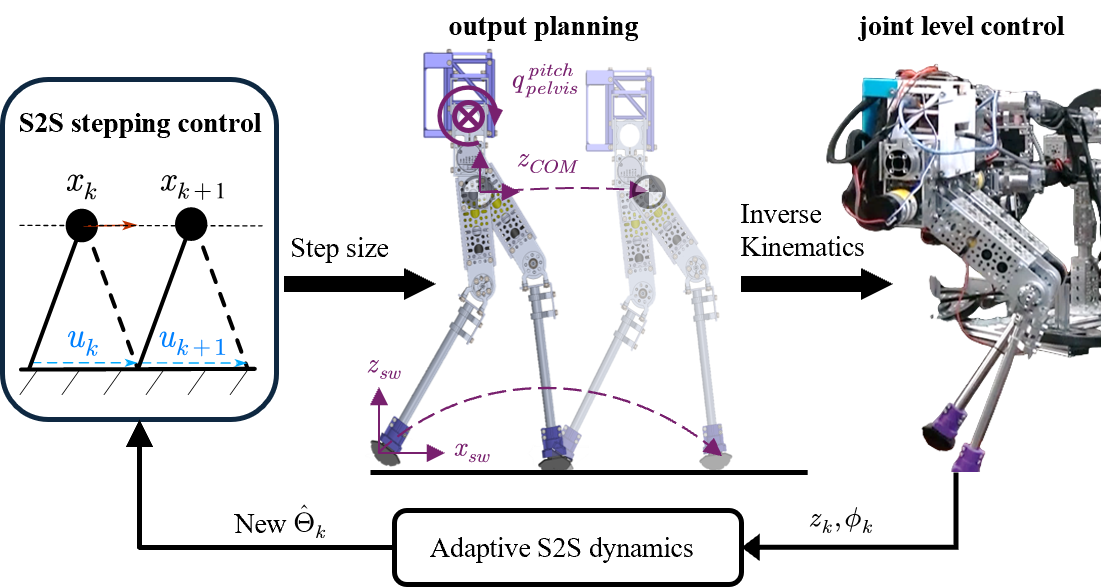}
    \caption{Structure of the walking controller Implementation based on S2S dynamics.}
    \vspace{-10pt}
    \label{hlip}
\end{figure}

\subsection{Hybrid Linear Inverted Pendulum Model}
 Bipedal walking can be naturally described by the step-to-step (S2S) dynamics \cite{S2S}. Mathematically, the S2S dynamics can be treated as a Poincar\'e return map at the Poincar\'e section which is normally chosen at the surface of the ground impact. The state of the robot is chosen as $x = [q,\dot{q}]^T$ where $q$ represents the generalized coordinates of the robot. Let $x_i^-$ denote the pre-impact state of the $i$-th ground impact during walking. The S2S dynamics of the robot can be written as $x_{k+1}^- = \mathcal{P}(x_k^-,\tau(t))$ where $\tau(t)$ is the control input over time.

Due to the highly nonlinear nature of the S2S dynamics of real bipedal robots, it is hard to acquire the analytical form of their S2S dynamics. Therefore, a linear approximation to the robot S2S dynamics is proposed in \cite{HLIP_TRO} which utilizes a reduced-order model called Hybrid Linear Inverted Pendulum (H-LIP). The H-LIP is a model with a constant center of mass (COM) height and two legs with point feet. Based on the number of feet in contact with the ground, the system is either in a single-stance phase (SSP) or a double-stance phase (DSP). The SSP phase is the same as the passive Linear Inverted Pendulum (LIP) model, and the DSP phase is assumed to have a constant horizontal COM velocity. Chosen the state of the H-LIP model as $x = [p,v]^T \in \mathbb{R}^2$, where $p$ and $v$ are the horizontal COM position and velocity w.r.t. the stance foot, respectively, the S2S dynamics of the H-LIP model then can be written as: 
\begin{equation} 
x_{k+1}^- = A^Hx_k^- +B^Hu_k,
\end{equation}
where $u_k\in \mathbb{R}$ represents the step size of the H-LIP model. The detailed derivation of the system matrix and input matrix can be found in \cite{HLIP_TRO}. The robot reduced-order S2S dynamics can be written as:
\begin{equation}
x_{k+1}^- = A^Hx_k^-+B^Hu_k +w_m,
\end{equation}
where $w_m = \mathcal{P}(x_k^-,\tau(t)) - A^Hx_k^- - B^Hu_k$ represents the integrated residual error between the S2S dynamics of the H-LIP model and the S2S dynamics of the real robot COM during one step. By constructing periodic walking gaits, $w_m$ lies in a bounded set $w_m\in W$ \cite{HLIP_TRO}. Then a state feedback controller called H-LIP based stepping controller can be designed to mitigate the error between the robot COM state and H-LIP state: $\bold{e} = x^R - x^{H}$. Taking $u^R = u_k^{H} + K(x_k^R-x_k^{H})$ yields the error S2S dynamics: 
\begin{equation}
\bold{e}_{k+1} = (A+BK)\bold{e}_k + w_k.
\end{equation}
The feedback $K$ can be designed through LQR or deadbeat controller to stabilize the system with $eig(A+BK)<1$.

The period orbits of the S2S dynamics of the robot can be characterized based on the number of steps completed during a walking period. Specifically Period-1 (P1) and Period-2 (P2) orbits are investigated. \\
\textbf{P1 Orbit:} The desired step size is uniquely determined by $u_d = v_dT$ where $v_d$ is the given desired walking velocity and $T$ is the desired step duration. Then the desired pre-impact state can be uniquely determined through the SSP dynamics of the H-LIP model and the desired step size.  \\
\textbf{P2 Orbit:} Different from the P1 orbit, the P2 orbit that can realize a desired velocity $v^d$ is not unique. The desired step sizes satisfy $u_L + u_R = 2v^dT$ where subscripts $_{L/R}$ denote the stance foot of that step, and $T$ is the step duration. In order to uniquely realize a P2 orbit, one should first choose a step size for the right stance or left stance foot. The desired states can then be uniquely determined for each step.  



\subsection{Adaptive Step-to-step Dynamics}
The H-LIP based model controller provides a baseline where the error of state tracking can be stabilized and bounded. To achieve higher accuracy and better performance, following the formulation in \cite{Ad_HLIP}, we leverage a static parametric model to reduce this modeling error; we then apply an adaptive controller to achieve better velocity tracking. 
To achieve this goal, we first review the robot model in the form of an adaptive control scheme. Generally, a static parametric model \cite{adaptive_tutor} is:
\begin{equation}
    z_k = \Theta^{*T}\phi_k,\end{equation}
where $z_k$ and $\phi_k$ are measurable signals, and $\Theta^*$ is the model we are trying to adapt to. According to the S2S dynamics of the H-LIP model developed above, we can expect a linear dynamics model with a constant offset:

\begin{equation}
\underbrace{x_k}_{z_k} =
\underbrace{(\begin{matrix} A&B&C\end{matrix})}_{\Theta^{*T}}
\underbrace{(\begin{matrix}x_{k-1}&u_{k-1}&1\end{matrix})^T}_{\phi_k},
\end{equation}
where $C$ represents the constant offset due to the nonlinear robot dynamics during stepping. Giving $z_k$ and $\phi_k$ at each step, we can use the update law:
\begin{equation}
    \hat{\Theta}_k = \hat{\Theta}_{k-1} + \Gamma \phi_{k} (\phi_{k}^T\phi_{k})^{-1}(z_k- \hat{\Theta}_{k-1}^T \phi_{k} ^T),
\end{equation}
where $\Gamma$ is a tunable gain for controlling the update size.
The performance of the adaptive controller will be evaluated in the experiment section to emphasize the capability of prototyping advanced control algorithms on STRIDE.

\subsection{Gait Design and Stepping on STRIDE} 
To apply the aforementioned H-LIP based approach on STRIDE, the designed gait should fulfill the requirement of the H-LIP model: the vertical COM position $z_{com}$ should be approximately constant w.r.t. the stance foot, the vertical position of the swing foot will be periodically lift-off and strike the ground, and the horizontal position of the swing foot will achieve the desired step size $u^d$ from the H-LIP based stepping controller. To fully constrain the walking, the orientation of the pelvis $q_{pelvis}^p$ is always controlled to certain desired values. Therefore the output during SSP is:
\begin{equation}
    \mathcal{Y} = \begin{bmatrix}z_{COM} \\ x_{sw} \\ z_{sw}\\ q_{pelvis}^{p} \end{bmatrix} - \begin{bmatrix}z_{COM}^d \\ x^d_{sw} \\ z^d_{sw}\\ q_{pelvis}^{p^d} \end{bmatrix}.    
\end{equation}
Since there are no compliant components in the mechanical design, and under the assumption of perfect plastic ground contact \cite{rigid_contact}, we assume only SSP will occur during walking. Therefore, only the trajectories of the SSP are designed.

The desired pitch angle of the pelvis is chosen to be constant. The rest of the output trajectories are designed with B\'ezier polynomials. The desired horizontal trajectory of the swing foot is designed as:
\begin{equation}
    x_{sw}^d = (1-b_h(t))x_{sw}^+ + b_h(t)u_x^d, 
\end{equation}
where $x_{sw}^+$ is the horizontal position of the swing foot w.r.t. the stance foot at the beginning of the current SSP, $u_x^d$ is the desired step size from the S2S based stepping controller, and $b_h(t)$ is a B\'ezier polynomial that transits from 0 ($t=0$) to 1 ($t = T_{ssp}$). The time variable $t$ will reset when the robot swaps its support leg. 
The desired vertical COM position is controlled to a constant $z_0$ which is the constant height of H-LIP. Due to a small jump when swapping the support leg, the desired trajectory should achieve a small transit from COM height after impact to the constant value. The desired trajectory is constructed as: 
\begin{equation} 
    z_{COM}^d = (1-b_h(t))z_{COM}^+ + b_h(t)z_0, 
\end{equation}
where $z_{COM}^+$ is the vertical position of the swing foot w.r.t the stance at the beginning of the current SSP, and $b_h(t)$ is the same B\'ezier polynomial aforementioned. 
The vertical position of the swing foot is controlled to perform periodic lift-off and touch-down behaviors. Another B\'ezier polynomial $b_v$ is used to design the trajectory in which the vertical position of the swing foot will first transit from $0 
\ (t=0)$ to $z_{sw}^{max}\ (t=\frac{T_{SSP}}{2})$ and then go back to $z_{sw}^{neg}\  (t=T_{SSP})$. $z_{sw}^{max}$ is a user-specified constant that determines the foot-ground clearance, and $z_{sw}^{neg}$ is a small negative number for making sure the foot will strike the ground at the end of SSP phase. The designed $z_{ws}^d$ is designed as:
\begin{equation}
    z_{sw}^d(t) = b_v(t,z_{sw}^{max}, z_{sw}^{neg}).  
\end{equation}

\subsection{Joint-level Controller Design}
Newton-Raphson inverse kinematics is employed to solve the joint trajectories from the desired output trajectories. The joint trajectories are then tracked by a PD plus feedforward controller. The feedforward torque is calculated via gravity compensation, and the desired torque is mapped to voltages that is sent to the motor drivers via PWM at 20kHz.

\blockUnderline{Gravity Compensation} The Euler-Lagrange dynamics of STRIDE during SSP is:
\begin{equation}
M(q)\ddot{q} + C(q,\dot{q}) + G(q) = Bu + J_s^TF_s,    
\end{equation}
where $M(q)$ is the mass matrix, $C(q,\dot{q})$ is the Coriolis and centrifugal forces, $G(q)$ is the gravity vector, $B$ is the input matrix, $J_s$ is the Jacobian matrix of the stance foot, $u$ is the control input and $F_s$ is the ground reaction force of that foot. Taking $\ddot{q} = \dot{q} = 0$, the proposed gravity compensation method can be formulated as a least-square optimization:
\begin{equation}
 \min_{u} \quad  ||G(q) - Bu -J_s^TF_s||^2,  
\end{equation}
which is solved using linear algebra functions in \texttt{Eigen}. 

\blockUnderline{Motor Control} Once the torques are solved, they are converted to a PWM value for motor control. The model of brushed DC motor with constant armature inductance described in \cite{Modern_robotics} is written as: 
\begin{align}
     V = k_e N \omega + IR,~
     \frac{\tau_m}{N} = k_tI,
\end{align}
where $V$, $I$ are the voltage supplied and current through the motor armatures, $R$ is the resistance of the armatures, $k_e$ and $k_t$ are the back EMF constant and motor torque constant, respectively, $N$ is the gear ratio, $\tau_m$ is the torque at output shaft and $\omega$ is the angular velocity of the motor at the output shaft. 
The torque at output shaft can then be mapped to the voltage by solving for $V$: \
\begin{equation}
     V =  k_e N \omega + \frac{R}{k_tN}\tau_m.
\end{equation}

\begin{figure}[t]
    \centering
    \includegraphics[width=1.0\linewidth]{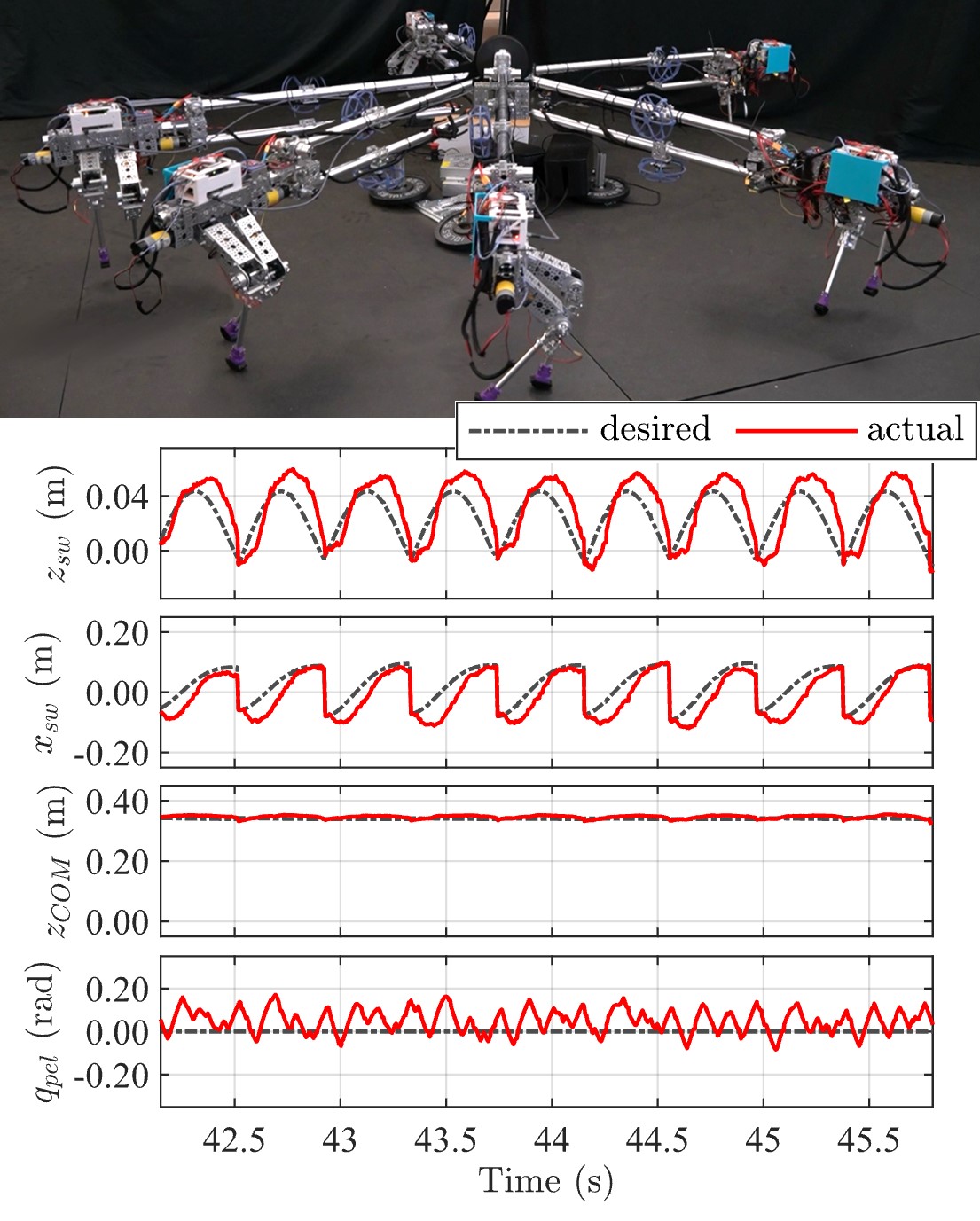}
    \caption{Stable walking with output tracking results.}
    \vspace{-10pt}
    \label{rigid}
\end{figure}

\section{System Demonstrations} \label{exp}

We then evaluate the performance of the proposed platform through several hardware experiments, with the controller developed in the previous section. Four experiments have been designed to analyze the potential of the platform: as an educational platform, as a testing platform for evaluating locomotion controls, and as a testing platform for design optimization. Fig. \ref{fig: mechanical design} (b) has depicted the experimental setup, and the experiment results can be seen in \cite{video}.

\subsection{Rigid Ground Walking}
 This experiment aims to demonstrate the capability of using the platform as an educational platform for legged locomotion. An H-LIP based S2S controller without adaptation is employed to synthesize desired walking behaviors. The walking height is chosen as $z_d = 0.34m$, and the desired walking speed is chosen as $v_d = 0.2m/s$. The desired walking periodic orbit is synthesized as a P1 orbit, with the walking period specified at $T = 0.3s$. Fig. \ref{rigid} illustrates the output trajectories of the robot. 
 
 Given that the hardware designs, electronic components, software, and walking controllers are open-sourced, students can reproduce the platform and the experiment following the provided detailed guidance. Learning outcomes can be analyzed based on the success of realizing walking behaviors. Moreover, the quality of the implementation can be evaluated based on the tracking errors between the desired and actual outputs. The errors currently come from two primary sources: (1) inverse kinematics error when mapping the desired outputs to desired joint outputs, which are relatively small, and (2) joint tracking errors, which are generally more apparent. A smaller error indicates better performance in the mechanical assembly, sensor filtering, and control implementation, reflecting the learning outcome of the students.

\subsection{Walking over Modular Challenging Soft Terrain}

\begin{figure}[t]
    \centering    
    \includegraphics[width=1\linewidth]{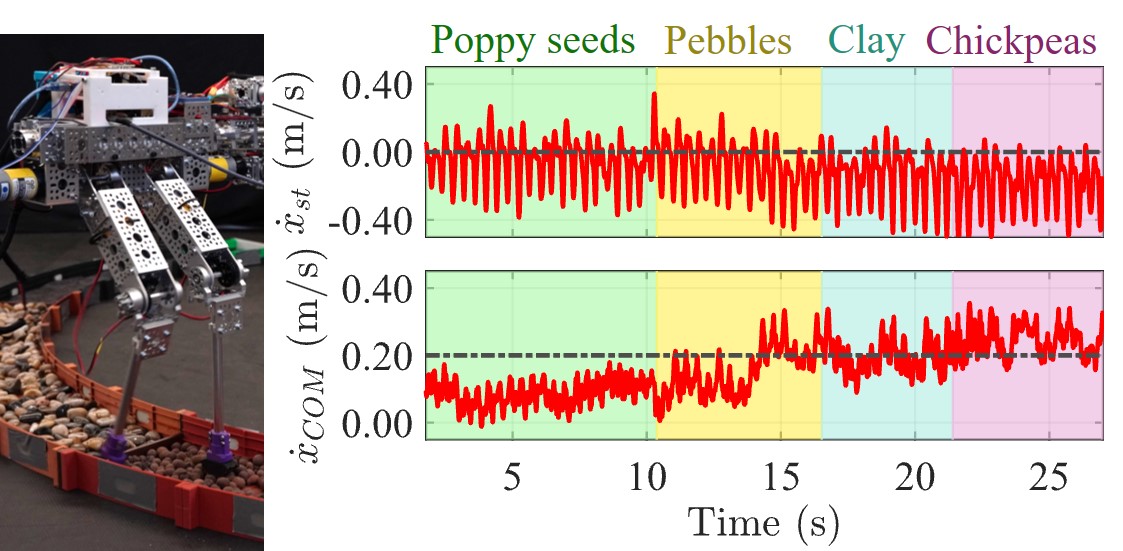}
    \caption{Rough terrain walking experiment.}
    \vspace{-0pt}
    \label{rough terrain}
\end{figure}

This experiment explores the potential of the platform for testing the robustness of different walking algorithms over rough terrains. Different terrains are used to simulate natural environments, making it easier to test real-world mobility indoors. The rough terrains are composed of poppy seeds, pebbles, clay pebbles, and chickpeas, respectively; they can be swapped to any other terrains easily. The walking parameters for this experiment are identical to those in the rigid ground walking experiment for comparative purposes. 

The stance foot and COM velocities in the horizontal direction are chosen as performance indicators. Fig. \ref{rough terrain} has depicted the walking performance of the robot on different terrains. Due to the varying surface roughness and hardness of the materials, the tracking behaviors of COM velocity and stance foot velocity of the robot differ. COM horizontal velocity of the robot is slower than expected when walking on poppy seeds and more jittery when walking on chickpeas and clay pebbles. The stance foot horizontal velocity of the robot is more slippery when walking on chickpeas and clay pebbles compared with walking on poppy seeds. The noticeable differences in velocity tracking across various terrains validate the effectiveness of the system in testing the robustness of walking algorithms on different surfaces.

\subsection{Quantitative Disturbance Injection}
\begin{figure}[t]
    \centering
    \includegraphics[width=1\linewidth]{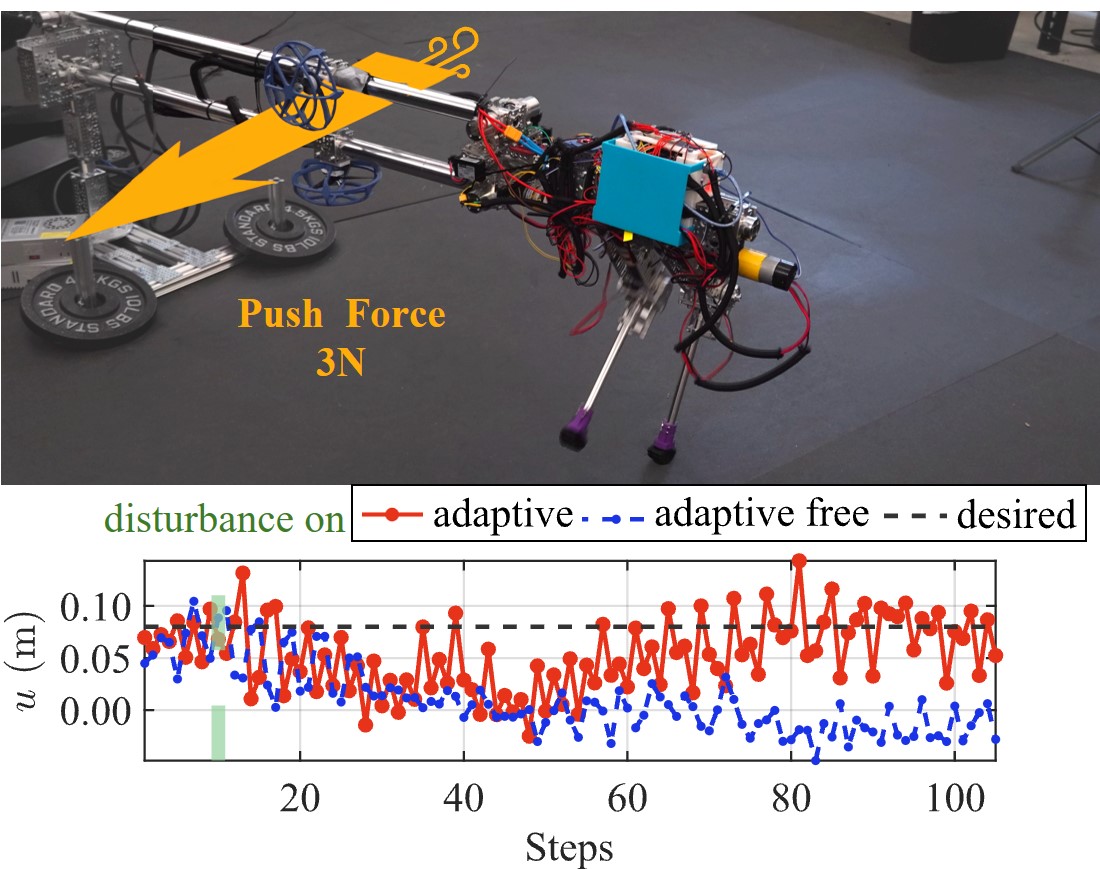}
    \caption{Adaptive control with disturbance experiment result.}
    \vspace{-0pt}
    \label{Adaptive control with disturbance experiment result}
\end{figure}

This experiment aims to provide the potential of applying the platform for advanced adaptive control validation and comparison. The walking parameter is chosen as the same as the rigid ground walking experiment, and the adaptive gain is chosen as $\Gamma = 0.4\mathtt{I}$ where $\mathtt{I}$ is the identity matrix. The horizontally mounted propeller provides a measurable disturbance of 3N.  Fig. \ref{Adaptive control with disturbance experiment result} has depicted the step tracking performance. 
Before the horizontal disturbance is applied, both controllers with or without adaptation have stable walking tracking. After the disturbance force is introduced, it can be easily seen that the adaptive controller provides better velocity tracking in terms of step sizes after a certain time of adaptation. The walking is stabilized to a mean step size of ~$0.08m$. A clear performance difference between the controller with and without adaptation has validated the use of this platform as a testing platform for the evaluation of advanced controls.  
\begin{figure}[t]
    \centering
    \includegraphics[width=1\linewidth]{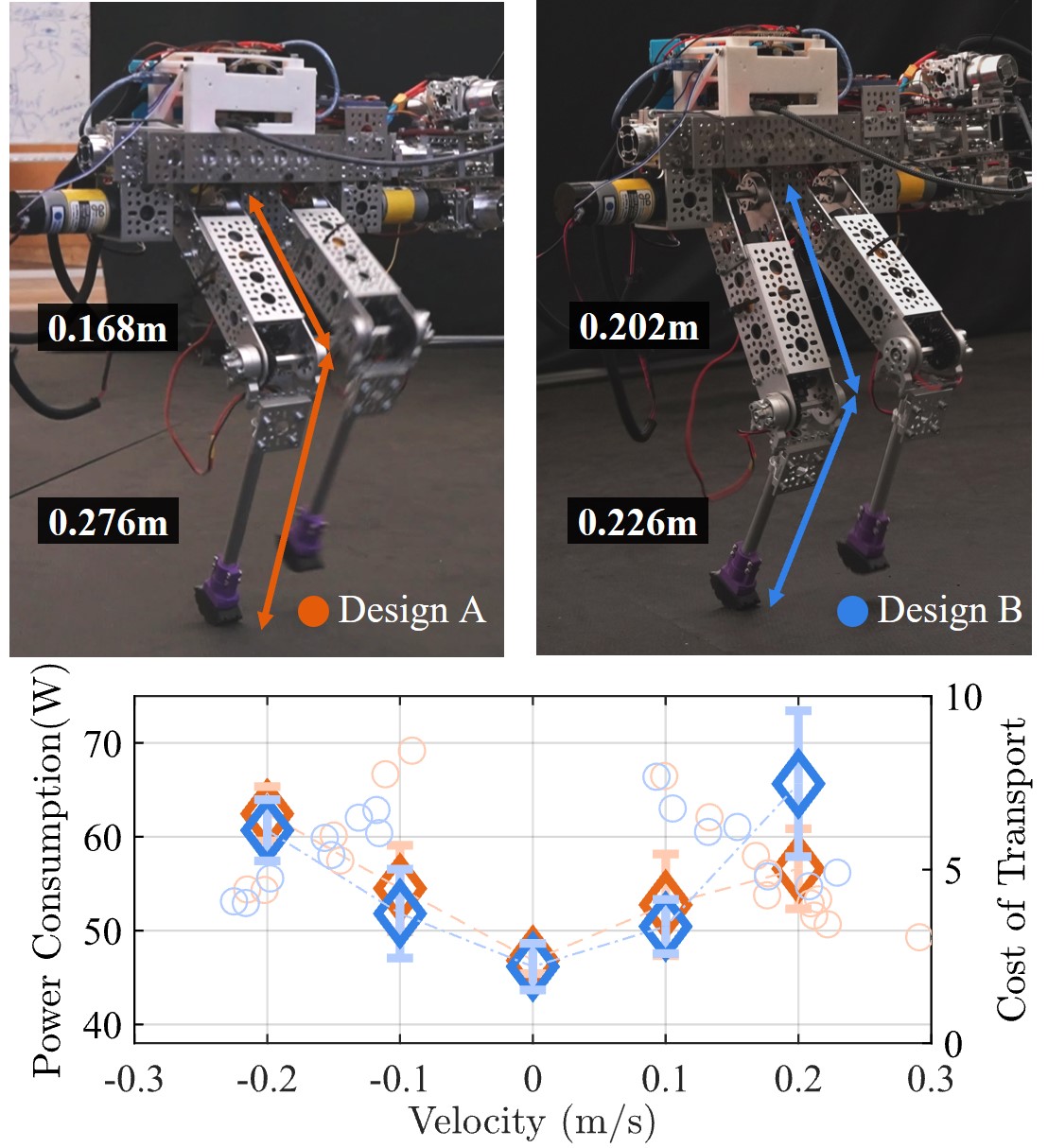}
    \caption{Power Consumption and Cost of Transport comparison on two different leg designs.}
    \vspace{-0pt}
    \label{COT&POW}
\end{figure}

\subsection{Design Evaluation Experiment}
In this experiment, we explore a potential scenario where design optimization is implemented to determine the optimal thigh-to-shin ratio. We implement two design candidates by changing the goBILDA parts shown in Fig. \ref{fig: mechanical design} (c). We then evaluate the two designs by comparing the power consumption and the cost of transport (COT) of walking, as shown in Fig. \ref{COT&POW}. The power consumption is lowest when the robot steps in place. Design A has relatively lower power consumption at higher positive velocities, while design B consumes less power at negative velocities. The COT for both designs is similar, with larger COT values occurring when the speed is smaller. As the speed increases, the COT decreases. This validates the purpose of this platform for rapid design evaluations on real hardware.

\section{Conclusion and Discussion}\label{conclu}
In conclusion, we present an open-source bipedal robotic platform that is low-cost, versatile, modular, robust, and customizable. The hardware system requires no machining, and it is primarily built from modular off-the-shelf parts for easy customization. The implemented controller using the S2S framework demonstrates the capability of deploying state-of-the-art walking controllers on the platform to locomote over natural terrains with push disturbances. Experiments validate its value for future education and research activities. 

In the future, we plan to use STRIDE to evaluate novel algorithms in motion planning, state estimation, feedback control, reinforcement learning, and bipedal locomotion to further demonstrate its value in research. In education, we plan to use STRIDE as a main platform in the robotics courses at UW-Madison, educating students on both robotics theories and engineering. We envision STRIDE will promote legged research and cultivate students with various expertise required in developing the next generations of dynamic, robust, and efficient legged robots.

\addtolength{\textheight}{.0cm}   






\bibliographystyle{IEEEtran}
\bibliography{reference}

\end{document}